\documentclass[11pt]{article}

\usepackage{arxiv}
\usepackage{amsmath,amssymb}
\usepackage{graphicx}
\usepackage{booktabs}
\usepackage{tabularx}
\usepackage{longtable}
\usepackage{multirow}
\usepackage{array}
\usepackage{makecell}
\usepackage[table]{xcolor}
\usepackage{enumitem}
\usepackage[font=small,labelfont=bf,skip=4pt]{caption}
\usepackage[round,authoryear]{natbib}
\usepackage{url}
\usepackage[colorlinks=true,linkcolor=blue!50!black,citecolor=blue!50!black,urlcolor=blue!50!black]{hyperref}
\usepackage{doi}

\graphicspath{{figures/}}
\setlist{itemsep=2pt,topsep=3pt}
\newcolumntype{L}[1]{>{\raggedright\arraybackslash}p{#1}}
\newcommand{\grp}[1]{\multicolumn{2}{@{}l}{\textbf{#1}}\\}

\renewcommand{\shorttitle}{Decision models for pathology data extraction}

\title{Can decision models extract clinical data from cancer pathology reports? An evaluation of accuracy, calibration, and failure modes}

\author{%
  Indranil Mallick\\
  Department of Radiation Oncology\\
  Tata Medical Center\\
  Kolkata, India
  \And
  Paromita Roy\\
  Department of Pathology\\
  Tata Medical Center\\
  Kolkata, India
}

\hypersetup{
  pdftitle={Can decision models extract clinical data from cancer pathology reports?},
  pdfauthor={Indranil Mallick, Paromita Roy},
  pdfkeywords={decision models, large language models, information extraction, pathology reports, oral cavity cancer, calibration},
}

\begin{document}
\maketitle

\begin{abstract}
Decision models are a new class of language model that, instead of generating text, choose one option per question from a fixed list and return a probability for every option. They have not yet been evaluated on clinical text. We evaluated two open-weights decision models, Nimble~9B and Tev1~4B, for extracting 25 fields from 1,113 oral cavity cancer pathology reports from a single centre, against a prospectively maintained clinical database. We compared them with two generative models: Qwen3.5~9B, run locally, and GLM-5.3-flash, run through a cloud API. Tev1 reached a mean field accuracy of 0.917 (95\% CI 0.913 to 0.920), the same as Qwen3.5~9B (0.918; $p = 0.26$), and was four times faster (15\,s against 61\,s per report). Nimble reached 0.862 and GLM-5.3-flash 0.986. Both decision models matched GLM-5.3-flash on simple dichotomous categories. They were weak on measurements binned into categories, on fields whose correct answer was not reported, and on some complex decisions (e.g.\ involved margins). We found that the main cause of Nimble's inferior performance was asking 25 questions in one request: with two questions per request, its accuracy improved. Pooled over fields, the returned probabilities were well calibrated (expected calibration error 0.022 and 0.026), but they did not flag systematic errors within single fields. A simulated strategy of sending answers with a probability below 0.94 to GLM-5.3-flash raised Tev1's accuracy to 0.982, with 38\% of answers escalated. Decision models can extract explicitly stated findings from pathology reports quickly on local hardware and return useful probabilities. They should be validated in the configuration in which they will be used, restricted to the fields in which they are reliable, and combined with other methods for numeric values.
\end{abstract}


\section{Introduction}

Pathology reports of oral cavity squamous cell carcinoma hold most of the information used to stage the disease and plan adjuvant treatment \citep{amin2017,roy2026}. Synoptic reporting to a standard protocol improves the completeness of this information \citep{seethala2026,sluijter2016}, but many reports are still written as narrative text. Turning them into the structured data needed for registries, audit and research requires manual abstraction by trained staff, which is slow, costly and itself a source of error.

Large language models (LLMs) can automate much of this work. Proprietary and open-weights generative models have extracted structured fields from pathology reports with high accuracy across cancer types \citep{grothey2025,truhn2024}, comparable to a human annotator in one study \citep{balasubramanian2025}, and models run on local hardware allow this without reports leaving the hospital \citep{leguellec2024,wiest2024}. Two problems remain. Generative models write their answers as text, which must be parsed and checked against the allowed values. They also give no direct measure of how likely each extracted value is to be correct: confidence stated in words is poorly calibrated \citep{boie2026}, and estimates built from token probabilities need extra computation or a separate calibration step \citep{gupta2026,schimmelpfennig2025}. Without such a measure, every extracted value has to be checked with the same care.

Decision models are a new class of language model, introduced in September 2026 \citep{jev}. A decision model receives a text ``state'' and a set of typed questions, each with a fixed list of options, and returns one option per question with a probability for every option, without generating text. They were built for fast, typed choices in software, such as routing support tickets, moderating content or playing games. Two open-weights decision models can be run locally: Nimble~9B, a low-rank adaptation of Qwen3.5~9B trained on 2,676 curated examples with synthetic labels \citep{nimble}, and Tev1~4B, an experimental fine-tune of Qwen3.5~4B that chooses among 2 to 24 options \citep{tev1}. Neither was trained on clinical text, and the Nimble developers note that its probabilities are normalised scores over the listed options, not calibrated estimates of correctness \citep{nimble}.

In principle, decision models suit two tasks in medical records. They could check data quality, by comparing a database entry with the answer read from the source report, and they could triage reports by the findings they contain, for example flagging reports with features that call for adjuvant treatment. Their probabilities could decide which answers to accept and which to send to a person or to a stronger model. All of this depends on how accurately the models identify key information in long, specialised reports unlike their training data, and we found no published evaluation of decision models on clinical text.

The aim of this study was to evaluate two open-weights decision models for extracting 25 fields from 1,113 oral cavity cancer pathology reports, against a prospectively maintained clinical database. We compared them with two generative LLMs: Qwen3.5~9B, the base model of Nimble, and GLM-5.3-flash, the most accurate model in a companion study on the same reports (manuscript in preparation). We measured accuracy, the calibration of the returned probabilities, internal consistency, robustness to option order and request size, and the effect of escalating low-probability answers to a stronger model or a person.

\section{Methods}

\subsection{Study design and reports}

This was a retrospective, single-centre study of structured data extraction from free-text surgical pathology reports. We used reports of patients with oral cavity squamous cell carcinoma treated with primary surgical resection at a tertiary care cancer hospital in India. The institutional ethics committee waived the requirement for informed consent.

Patients were treated between 2012 and 2020. Reports were de-identified before analysis. Of 1,194 reports, 81 were excluded: 58 because the report stated fewer than 5 of the 25 study factors and 23 because the primary tumour was not in the oral cavity. The remaining 1,113 reports formed the scored cohort. Their median length was 2,102 characters (interquartile range 1,773 to 2,567; range 354 to 4,361).

\subsection{Reference standard}

The reference standard was a prospectively maintained clinical database of the same patients. Values were entered during routine care and checked against the reports by an expert clinician and a pathologist. No model output was used to build or correct the reference standard. Database values were mapped to the 25 study fields, and numeric values were converted to categories as described below.

\subsection{Extraction schema}

The schema had 25 fields aligned to the oral cavity dataset, the CAP oral cavity protocol and AJCC 8th edition staging \citep{amin2017,seethala2026}. Sixteen were categorical: primary subsite, laterality, focality, histological type, grade, perineural invasion (PNI), lymphovascular invasion (LVI), worst pattern of invasion~5 (WPOI-5), bone invasion, masticator space involvement, skin involvement, margin status, side of neck dissection, right and left neck node involvement, and extranodal extension (ENE). Three recorded the pathological T category, N category and AJCC edition as transcribed from the report, not derived. The six numeric fields were scored as categories (Table~\ref{tab:bins}), because decision models can only choose among options. A value on a cut-off was assigned to the lower category. For reporting, fields were grouped as primary tumour (15 fields), nodal (7), staging (3) and binned numeric (6).

\begin{table}[htbp]
\centering
\caption{Numeric fields and the categories used for scoring.}
\label{tab:bins}
\small
\begin{tabular}{@{}ll@{}}
\toprule
Field & Categories \\
\midrule
Tumour size (greatest dimension) & $\le$20\,mm; $>$20 to 40\,mm; $>$40\,mm \\
Depth of invasion & $\le$5\,mm; $>$5 to 10\,mm; $>$10\,mm \\
Tumour thickness & $\le$5\,mm; $>$5 to 10\,mm; $>$10\,mm \\
Number of positive nodes & 0; 1; 2 or more; no neck dissection \\
Number of nodes examined & 1 to 17; 18 or more; no neck dissection \\
Largest involved node & $\le$30\,mm; $>$30 to 60\,mm; $>$60\,mm; not applicable \\
 & \quad (no positive nodes or no neck dissection) \\
\bottomrule
\end{tabular}
\end{table}

In every field a blank value meant ``not stated in the report'' and was treated as a valid label.

\subsection{Decision models}

We evaluated two open-weights decision models available for local use: Nimble~9B (Bespoke Labs; Qwen3.5-9B with a LoRA adapter; Ollama tag \texttt{nimble:9b}, Q8\_0) and Tev1~4B (Together AI, released as experimental; tag \texttt{tev1:4b}, model digests in Table~\ref{tab:questions}). Both were served by Ollama 0.40.6 and called from a Python client that uses only the standard library. Decision models do not generate text. They receive a text ``state'' and a set of typed questions, and return one chosen option per question with a probability for every option.

The state was the report text, preceded by a two-sentence description of the report type. The 25 questions were all of the ``choice'' type, with an instruction and a one-line description of each option. Question wording and option labels were taken from the field definitions given to the generative models, and a script checked that every option matched a value allowed in the generative schema. All 25 questions were sent in one request per report. The endpoint builds a separate prompt for each question from the state and that question with its options, so the report is processed once per question. The endpoint resolves ties in favour of the first listed option. The questions were not revised after seeing results.

All local runs used a machine with an Intel Arc GPU (Intel Core Ultra~7, 32\,GB RAM).

\subsection{Generative comparators}

Two generative LLMs served as comparators. Qwen3.5~9B ran locally on the same machine; it is the base model of Nimble and close in size to Nimble and about twice the size of Tev1. GLM-5.3-flash ran through the Ollama Cloud API and was the most accurate generative model in a companion study on the same reports (manuscript in preparation). All generative models used one prompt file containing a system prompt, a user template and a definition for every field. Each report was sent in one request, and output was constrained to a JSON schema listing the allowed values. Temperature was 0 and the context window 8,192 tokens per request. Thinking was turned off for local models but was retained for the cloud models. Each model was run once on 8--11 October 2026. The numeric outputs of the generative models were binned with the same function as the reference standard, so all models were scored on identical categories.

\subsection{Scoring}

All 1,113 reports were scored for every model. A missing report or a failed request counted as an all-blank answer. None occurred for any of the four models. Answers were compared with the reference by exact match after normalising case and whitespace. A blank reference value matched a blank answer. The decision models' ``Not reported'' option, and ``None'' for neck dissection side, were treated as blank, except for skin, where ``Not reported'' was itself a reference label. Decision models had no blank option for focality, histological type, bone invasion or masticator space; a sensitivity analysis merged blank with each field's stand-in option.

\subsection{Statistical analysis}

The primary outcome was mean field accuracy (MFA): the proportion of correct report-field cells over all 25 fields. Its 95\% confidence interval came from a case-level percentile bootstrap (2,000 resamples). Differences in MFA between pairs of models were estimated with a paired case-level bootstrap, overall and by field group, with a two-sided bootstrap $p$ value. Per-field accuracy was reported with Wilson 95\% intervals and compared between models with exact McNemar tests, Holm-corrected across the 25 fields within each pair. We also report Cohen's kappa (blank as its own category), linear-weighted kappa for ordinal fields, and the sensitivity, specificity, positive and negative predictive values of 14 clinically decisive findings.

For the decision models, calibration was assessed with the probability each model gave to its chosen option. We report the area under the ROC curve (AUROC) for the probability as a predictor of a correct answer, the expected calibration error (10 equal-width bins), the Brier score, and the share and accuracy of answers above probability thresholds of 0.8, 0.9 and 0.95.

Internal consistency was measured with nine rules that a single report cannot break, such as pN0 with positive nodes, or cortical bone invasion with pT1 to T3 (Table~\ref{tab:rules}). The rules were applied to each model's own answers and to the reference standard; the reference rate shows how often real reports break a rule.

A cascade was simulated in which a decision-model answer was kept when its probability was at least a threshold $t$, and otherwise replaced by the answer of another model (Qwen3.5~9B or GLM-5.3-flash) or by a human reviewer assumed to be correct.

Two robustness experiments were run on the decision models. In the first, the T category and margin status questions were asked with the options in the original order, reversed, in a fixed random order, and (for T category only; see Results) without Tx, T0 and Tis, on a seeded random sample of 200 reports. Only these two questions were sent, so all variants had the same context. In the second, all 25 questions were asked again for a seeded sample of 100 reports to check that repeated runs give identical answers. Because the option-order runs sent two questions per request and the main run sent 25, the original-order arm also allowed a post hoc comparison of the two request sizes on the same reports. Paired comparisons used exact McNemar tests. Analyses used Python with numpy, pandas and matplotlib.

\section{Results}

\subsection{Cohort and reference labels}

The 1,113 scored reports covered oral tongue (501, 45.0\%), buccal mucosa (362, 32.5\%), lower alveolus (144, 12.9\%), retromolar trigone (34), hard palate (15), floor of mouth (15), upper alveolus (15), lip (14) and overlapping tumours (13). Almost all patients had a neck dissection (1,101). Several fields were often not stated in the report: thickness in 711 reports, WPOI in 697, AJCC edition in 702, depth of invasion in 370 and T category in 118 (Table~\ref{tab:labels}).

\begin{table}[p]
\centering
\caption{Reference labels in the scored reports ($n = 1{,}113$).}
\label{tab:labels}
\footnotesize
\renewcommand{\arraystretch}{0.92}
\begin{minipage}[t]{0.48\textwidth}
\centering
\begin{tabular}[t]{@{}lr@{}}
\toprule
Field and label & $n$ (\%) \\
\midrule
\grp{Grade}
Moderately differentiated & 648 (58.2\%) \\
Poorly differentiated & 274 (24.6\%) \\
Well differentiated & 111 (10.0\%) \\
Not stated & 80 (7.2\%) \\
\grp{PNI}
Absent & 621 (55.8\%) \\
Present & 464 (41.7\%) \\
Indeterminate & 2 (0.2\%) \\
Not stated & 26 (2.3\%) \\
\grp{LVI}
Absent & 709 (63.7\%) \\
Present & 336 (30.2\%) \\
Indeterminate & 44 (4.0\%) \\
Not stated & 24 (2.2\%) \\
\grp{WPOI}
WPOI-5 present & 125 (11.2\%) \\
WPOI-5 and WPOI-4 present & 2 (0.2\%) \\
WPOI-4 present, WPOI-5 absent & 25 (2.2\%) \\
WPOI-5 absent & 264 (23.7\%) \\
Not stated & 697 (62.6\%) \\
\grp{Bone}
Not resected & 532 (47.8\%) \\
Not involved & 279 (25.1\%) \\
Cortical invasion & 239 (21.5\%) \\
Erosion & 49 (4.4\%) \\
Not stated & 14 (1.3\%) \\
\grp{Masticator space}
Not resected & 771 (69.3\%) \\
Absent & 275 (24.7\%) \\
Present & 63 (5.7\%) \\
Not stated & 4 (0.4\%) \\
\grp{Skin}
Not reported & 933 (83.8\%) \\
Absent & 126 (11.3\%) \\
Present & 54 (4.9\%) \\
\grp{Margin}
Clear & 899 (80.8\%) \\
Close ($<$5\,mm) & 174 (15.6\%) \\
Involved & 33 (3.0\%) \\
Not stated & 7 (0.6\%) \\
\grp{Focality}
1 focus & 1,088 (97.8\%) \\
2 foci & 22 (2.0\%) \\
3 or more foci & 2 (0.2\%) \\
Not stated & 1 (0.1\%) \\
\grp{Histological type}
Conventional SCC & 1,056 (94.9\%) \\
Verrucous & 14 (1.3\%) \\
Spindle cell & 14 (1.3\%) \\
Papillary & 3 (0.3\%) \\
Adenosquamous & 1 (0.1\%) \\
Other & 5 (0.4\%) \\
Not stated & 20 (1.8\%) \\
\bottomrule
\end{tabular}
\end{minipage}\hfill
\begin{minipage}[t]{0.48\textwidth}
\centering
\begin{tabular}[t]{@{}lr@{}}
\toprule
Field and label & $n$ (\%) \\
\midrule
\grp{Tumour size}
$\le$20\,mm & 298 (26.8\%) \\
$>$20 to 40\,mm & 527 (47.3\%) \\
$>$40\,mm & 248 (22.3\%) \\
Not stated & 40 (3.6\%) \\
\grp{Depth of invasion}
$\le$5\,mm & 183 (16.4\%) \\
$>$5 to 10\,mm & 233 (20.9\%) \\
$>$10\,mm & 327 (29.4\%) \\
Not stated & 370 (33.2\%) \\
\grp{Thickness}
$\le$5\,mm & 96 (8.6\%) \\
$>$5 to 10\,mm & 121 (10.9\%) \\
$>$10\,mm & 185 (16.6\%) \\
Not stated & 711 (63.9\%) \\
\grp{Positive nodes}
0 & 634 (57.0\%) \\
1 & 186 (16.7\%) \\
2 or more & 281 (25.2\%) \\
No neck dissection & 11 (1.0\%) \\
Not stated & 1 (0.1\%) \\
\grp{Nodes examined}
1 to 17 & 28 (2.5\%) \\
18 or more & 1,073 (96.4\%) \\
No neck dissection & 11 (1.0\%) \\
Not stated & 1 (0.1\%) \\
\grp{Largest involved node}
$\le$30\,mm & 344 (30.9\%) \\
$>$30 to 60\,mm & 30 (2.7\%) \\
$>$60\,mm & 1 (0.1\%) \\
Not applicable & 645 (58.0\%) \\
Not stated & 93 (8.4\%) \\
\grp{ENE}
Yes & 239 (21.5\%) \\
No & 874 (78.5\%) \\
\grp{pT}
T1 & 211 (19.0\%) \\
T2 & 332 (29.8\%) \\
T3 & 164 (14.7\%) \\
T4a & 235 (21.1\%) \\
T4b & 50 (4.5\%) \\
Tis & 1 (0.1\%) \\
Tx & 2 (0.2\%) \\
Not stated & 118 (10.6\%) \\
\grp{pN}
N0 & 562 (50.5\%) \\
N1 & 160 (14.4\%) \\
N2a & 15 (1.3\%) \\
N2b & 150 (13.5\%) \\
N2c & 24 (2.2\%) \\
N3b & 78 (7.0\%) \\
Nx & 8 (0.7\%) \\
Not stated & 116 (10.4\%) \\
\grp{AJCC edition}
8th & 407 (36.6\%) \\
7th & 4 (0.4\%) \\
Not stated & 702 (63.1\%) \\
\bottomrule
\end{tabular}
\end{minipage}
\end{table}

\subsection{Overall accuracy}

\begin{figure}[htbp]
\centering
\includegraphics[width=0.77\textwidth]{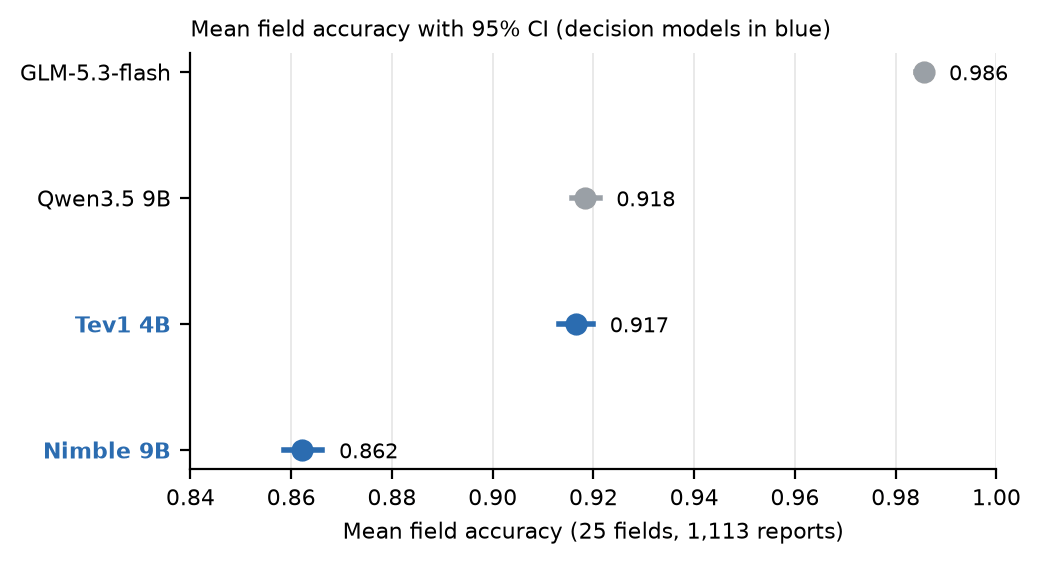}
\caption{Mean field accuracy with 95\% confidence intervals (case-level bootstrap, 2,000 resamples) for the two decision models (blue) and the two generative comparators; 25 fields, 1,113 reports.}
\label{fig:mfa}
\end{figure}

Tev1~4B reached an MFA of 0.917 (95\% CI 0.913 to 0.920), the same as the generative Qwen3.5~9B (0.918; difference $-0.002$, 95\% CI $-0.005$ to 0.001, $p = 0.26$). Nimble~9B reached 0.862 (0.858 to 0.866), 0.056 below its own base model Qwen3.5~9B (95\% CI 0.053 to 0.060) and 0.054 below Tev1 (0.050 to 0.058). GLM-5.3-flash reached 0.986 (0.984 to 0.987), 0.069 above Tev1 (95\% CI 0.066 to 0.072) and 0.124 above Nimble (0.120 to 0.127) (Table~\ref{tab:accuracy}, Figure~\ref{fig:mfa}). Merging blank with the stand-in options changed MFA by about 0.0002.

Although Tev1 and Qwen3.5~9B had the same overall accuracy, their errors fell in different fields. Tev1 was more accurate on primary tumour fields (difference 0.018, 95\% CI 0.014 to 0.022) and less accurate on nodal fields ($-0.044$, $-0.051$ to $-0.038$) and binned numeric fields ($-0.103$, $-0.112$ to $-0.095$); staging fields did not differ ($-0.002$, $-0.008$ to 0.004).

\begin{table}[htbp]
\centering
\caption{Accuracy of the decision models and the two generative comparators. Group columns give accuracy over the fields in each group. Time is the mean per report; cloud times are not comparable with local times.}
\label{tab:accuracy}
\footnotesize
\setlength{\tabcolsep}{4pt}
\begin{tabular}{@{}llcccccccr@{}}
\toprule
 & & & \multicolumn{4}{c}{Accuracy by field group} & & Reports & \\
\cmidrule(lr){4-7}
Model & Type & MFA (95\% CI) & \makecell{Primary\\tumour} & Nodal & Staging & \makecell{Binned\\numeric} & \makecell{Mean\\kappa} & \makecell{fully\\correct} & \makecell{Time\\(s)} \\
\midrule
GLM-5.3-flash & Generative, cloud & 0.986 (0.984--0.987) & 0.984 & 0.983 & 0.998 & 0.985 & 0.951 & 73.3\% & 23.6 \\
Qwen3.5 9B & Generative, local & 0.918 (0.916--0.921) & 0.887 & 0.971 & 0.953 & 0.971 & 0.822 & 5.7\% & 60.8 \\
\textbf{Tev1 4B} & Decision, local & 0.917 (0.913--0.920) & 0.905 & 0.927 & 0.951 & 0.868 & 0.810 & 10.0\% & 15.0 \\
\textbf{Nimble 9B} & Decision, local & 0.862 (0.858--0.866) & 0.834 & 0.962 & 0.769 & 0.871 & 0.740 & 1.8\% & 44.3 \\
\bottomrule
\end{tabular}
\end{table}

\subsection{Field-level accuracy}

\begin{figure}[htbp]
\centering
\includegraphics[width=0.85\textwidth]{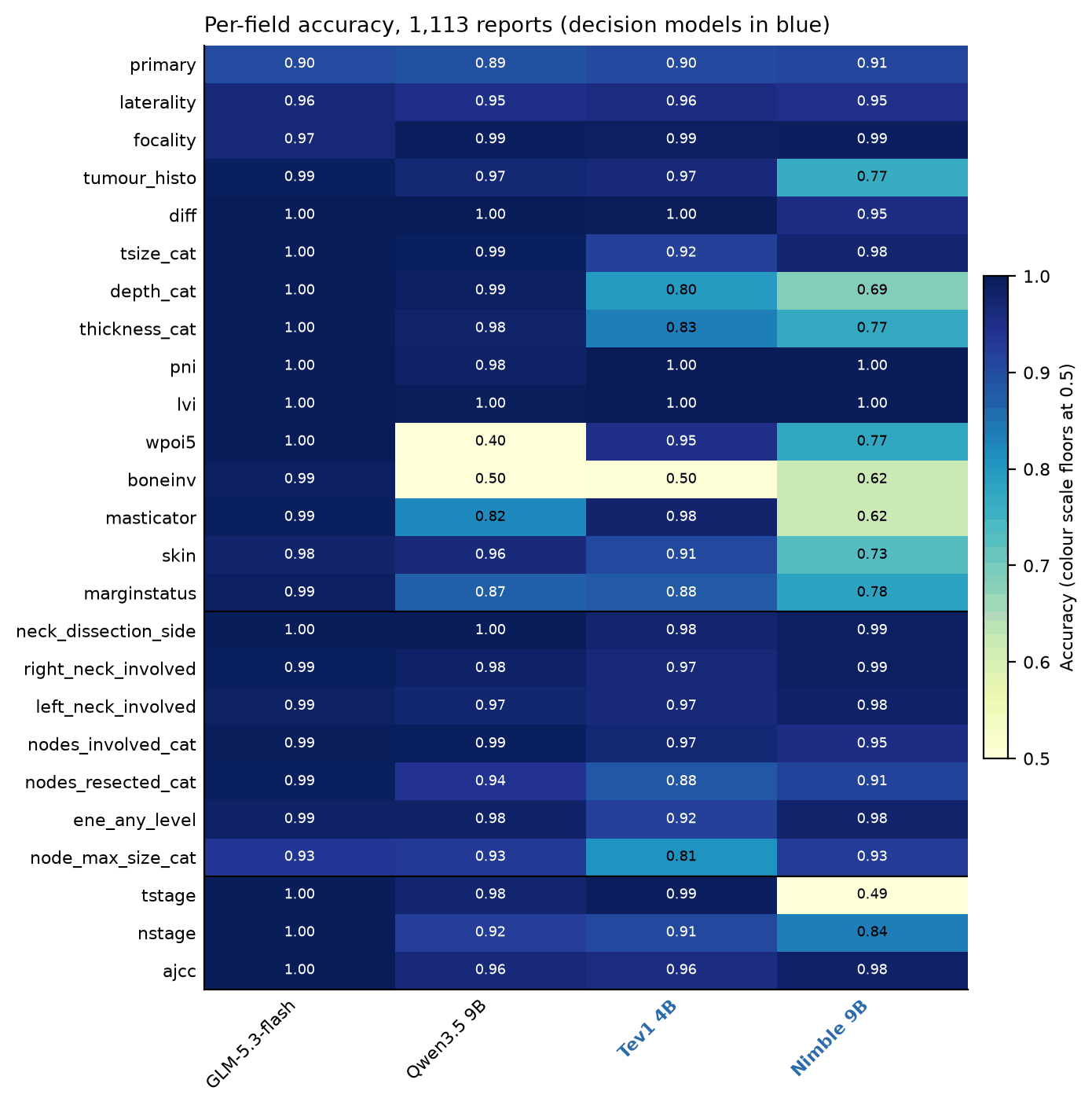}
\caption{Accuracy by field for the two decision models (blue) and the two generative comparators. The colour scale floors at 0.5.}
\label{fig:fields}
\end{figure}

Both decision models matched GLM-5.3-flash on PNI (Tev1 1.000, Nimble 0.997), LVI (0.998, 0.997), focality (0.990, 0.991) and laterality (0.957, 0.946), and Tev1 also on grade (0.996) and pT (0.992) (Figure~\ref{fig:fields}). Against Qwen3.5~9B, Tev1 was more accurate on WPOI-5 (0.949 vs 0.399), masticator space (0.978 vs 0.821), PNI and pT, and less accurate on tumour size, depth of invasion, thickness, largest node size, nodes examined, ENE and skin (all Holm-corrected $p < 0.05$).

\begin{figure}[htbp]
\centering
\includegraphics[width=\textwidth]{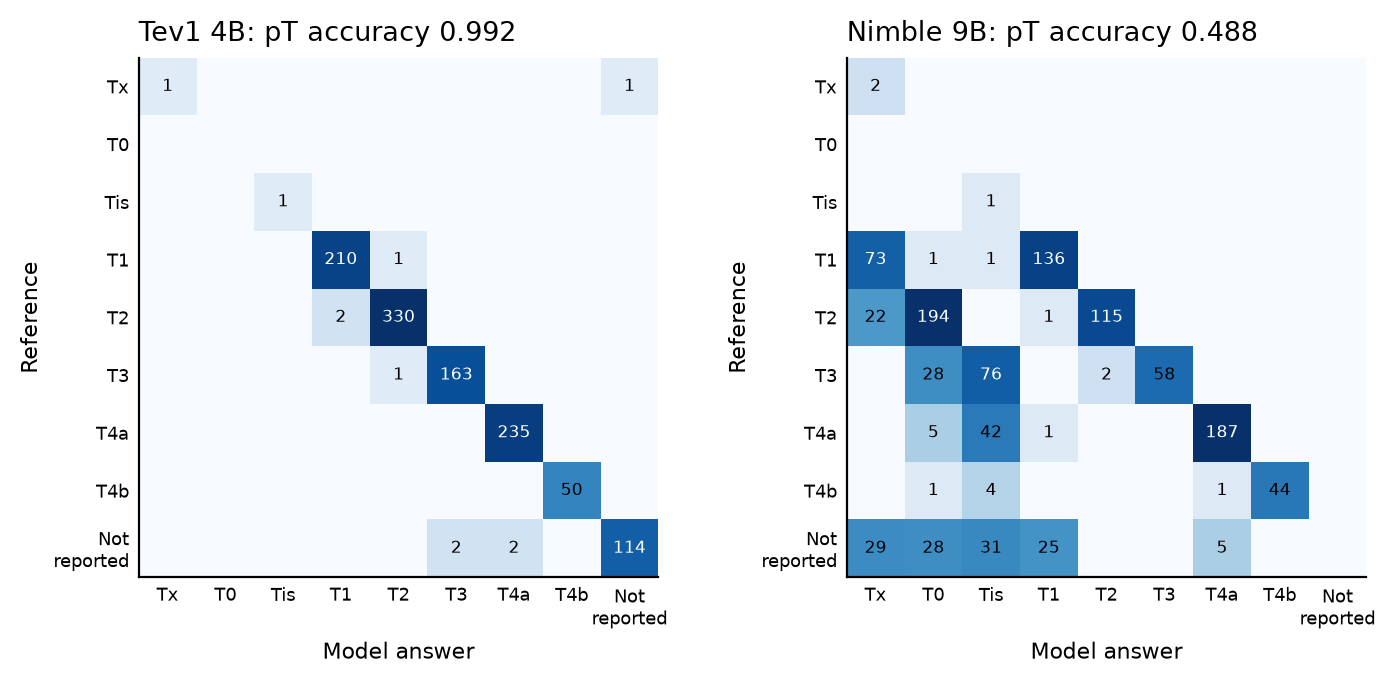}
\caption{Reference pT against model answer for Tev1~4B (left) and Nimble~9B (right) when all 25 questions were asked in one request. Nimble's answers for T1 to T3 tumours fall largely on Tx, T0 and Tis.}
\label{fig:pt}
\end{figure}

The decision models' errors were concentrated in a few fields, and most followed a pattern rather than looking like random misreading:
\begin{itemize}
  \item Nimble's pT accuracy was 0.488 against 0.992 for Tev1 on the same question. Nimble answered T0 for 194 of 332 T2 tumours, Tx for 73 of 211 T1 tumours and Tis for 76 of 164 T3 tumours, while T4a was mostly correct (187 of 235) (Figure~\ref{fig:pt}). The robustness experiments showed that this error came from asking all 25 questions in one request, not from the order of the options (see Section~\ref{sec:robustness}).
  \item Tev1 never chose ``Bone not resected'', so the 532 reports without bone in the specimen were labelled ``Not involved'' (bone accuracy 0.503).
  \item Nimble called basaloid histology in 179 reports, against none in the reference, and called the margin close or involved in 424 reports against 207.
  \item Both decision models were least accurate on depth of invasion (Tev1 0.795, Nimble 0.686) and thickness (0.835, 0.770), where GLM-5.3-flash and Qwen3.5~9B both scored 0.980 or more.
\end{itemize}

\subsection{Clinically decisive findings}

Sensitivity for PNI, LVI, WPOI-5 and cortical bone invasion was 0.98 to 1.00 for all four models (Table~\ref{tab:findings}). The decision models differed from the generative models on margins and measurements. Tev1 detected 16 of 33 involved margins (sensitivity 0.48) and about half of tumours with depth of invasion over 10\,mm (0.51). Nimble over-called close or involved margins (positive predictive value 0.47) and masticator space involvement (0.62).

\begin{table}[htbp]
\centering
\caption{Sensitivity / specificity for selected findings. $n$ = reference positives.}
\label{tab:findings}
\small
\begin{tabular}{@{}lrcccc@{}}
\toprule
Finding & $n$ & Tev1 4B & Nimble 9B & Qwen3.5 9B & GLM-5.3-flash \\
\midrule
PNI present & 464 & 1.000 / 1.000 & 1.000 / 0.998 & 1.000 / 0.997 & 1.000 / 0.998 \\
LVI present & 336 & 0.997 / 1.000 & 1.000 / 0.997 & 1.000 / 0.997 & 1.000 / 0.997 \\
WPOI-5 present & 127 & 1.000 / 0.999 & 1.000 / 0.995 & 1.000 / 1.000 & 1.000 / 1.000 \\
Cortical bone invasion & 239 & 0.983 / 0.994 & 0.987 / 0.999 & 0.983 / 0.994 & 0.996 / 1.000 \\
Masticator space involved & 63 & 0.952 / 0.995 & 1.000 / 0.964 & 0.968 / 0.990 & 1.000 / 0.998 \\
Skin involved & 54 & 1.000 / 0.983 & 0.981 / 0.990 & 1.000 / 0.985 & 1.000 / 0.995 \\
Margin involved & 33 & 0.485 / 0.997 & 0.758 / 0.994 & 0.697 / 0.997 & 0.939 / 0.999 \\
Margin close or involved & 207 & 0.657 / 0.939 & 0.961 / 0.752 & 0.348 / 0.996 & 0.990 / 0.990 \\
Tumour size $>$40\,mm & 248 & 0.702 / 0.999 & 0.968 / 0.997 & 1.000 / 0.999 & 1.000 / 1.000 \\
Depth of invasion $>$10\,mm & 327 & 0.511 / 0.990 & 0.838 / 0.879 & 1.000 / 0.997 & 0.997 / 1.000 \\
ENE present & 239 & 0.929 / 0.999 & 0.992 / 0.999 & 0.987 / 0.999 & 0.996 / 0.999 \\
2 or more positive nodes & 281 & 0.947 / 0.999 & 0.868 / 1.000 & 0.989 / 0.999 & 0.989 / 0.999 \\
pT4 & 285 & 1.000 / 0.998 & 0.814 / 0.994 & 1.000 / 0.998 & 1.000 / 1.000 \\
pN+ & 427 & 1.000 / 0.959 & 0.850 / 0.968 & 1.000 / 0.991 & 1.000 / 0.999 \\
\bottomrule
\end{tabular}
\end{table}

\subsection{Calibration and selective acceptance}

\begin{figure}[htbp]
\centering
\includegraphics[width=\textwidth]{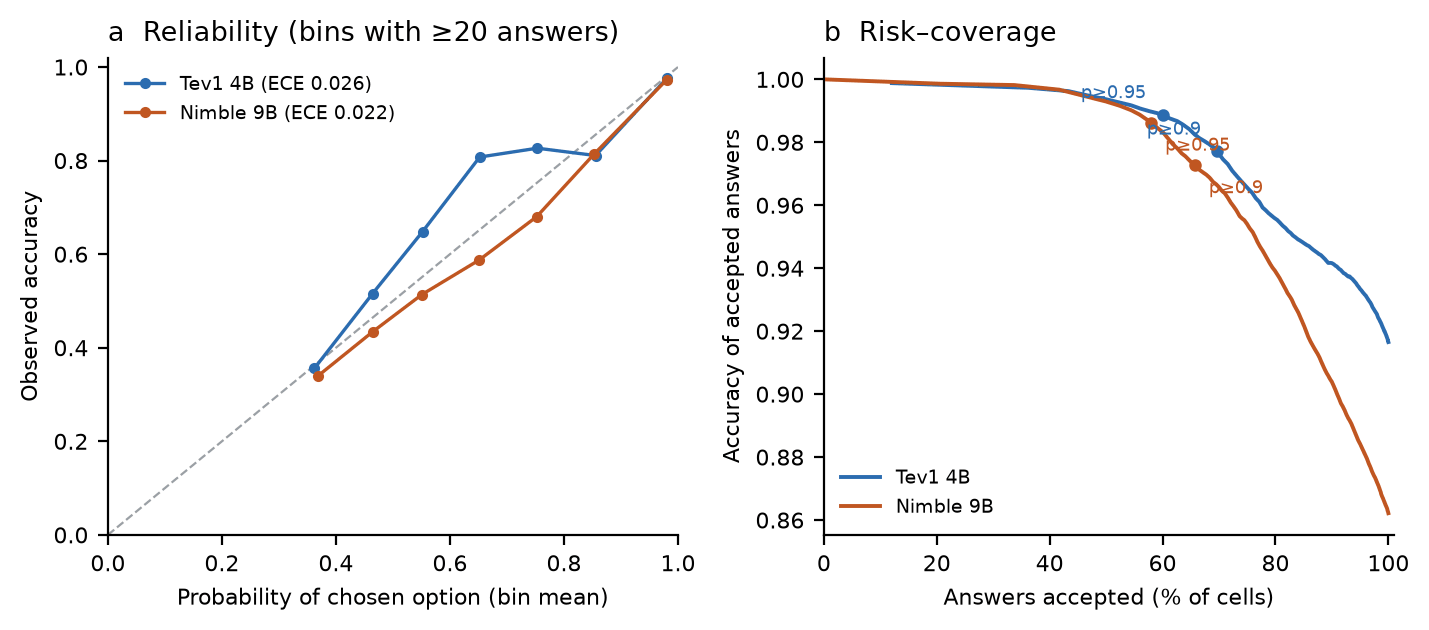}
\caption{(a) Reliability diagram: observed accuracy against the probability of the chosen option, pooled over all fields (bins with at least 20 answers). (b) Accuracy of the answers kept against the share of answers kept, as the probability threshold for acceptance rises.}
\label{fig:calib}
\end{figure}

Pooled over all fields, the probability of the chosen option predicted whether the answer was correct, with an AUROC of 0.878 for Nimble and 0.844 for Tev1. Expected calibration error was 0.022 and 0.026, and the Brier score 0.089 and 0.067 (Figure~\ref{fig:calib}a). Accepting only answers with probability of at least 0.9 kept 65.7\% of Nimble's answers at 97.3\% accuracy and 69.7\% of Tev1's at 97.7\%. At 0.95, the figures were 58.0\% at 98.6\% and 60.0\% at 98.9\% (Figure~\ref{fig:calib}b). Calibration was poorest in the fields with systematic errors: Tev1 bone invasion (expected calibration error 0.370; mean probability 0.845 at an accuracy of 0.503), Tev1 nodes examined (0.210), Nimble bone invasion (0.192) and Nimble pT (0.187).

\subsection{Internal consistency}

Reports with at least one contradiction between linked fields numbered 23 in the reference standard, 22 for GLM-5.3-flash, 66 for Qwen3.5~9B, 91 for Tev1 and 112 for Nimble (Table~\ref{tab:consistency}). Tev1's excess came mostly from the side of neck dissection disagreeing with right or left node status (42 reports). Nimble's came from the node count disagreeing with pN (67), with right or left node status (50), with the largest node size (47) and with ENE (34). Because Nimble's pT errors moved answers to Tx, T0 and Tis, which the pT rules do not test, its pT contradictions are undercounted.

\begin{table}[htbp]
\centering
\caption{Reports breaking each consistency rule (definitions in Table~\ref{tab:rules}).}
\label{tab:consistency}
\small
\begin{tabular}{@{}L{6.4cm}rrrrr@{}}
\toprule
Rule & Reference & GLM-5.3-flash & Qwen3.5 9B & Tev1 4B & Nimble 9B \\
\midrule
Neck dissection side vs right/left node status & 3 & 0 & 4 & 42 & 11 \\
Positive-node count vs right/left node status & 6 & 2 & 17 & 13 & 50 \\
Positive-node count vs pN & 4 & 8 & 10 & 23 & 67 \\
ENE with no positive nodes & 0 & 0 & 0 & 1 & 34 \\
Largest node size vs node count & 0 & 0 & 0 & 5 & 47 \\
``No neck dissection'' vs other node fields & 1 & 1 & 4 & 0 & 5 \\
Cortical bone, skin or masticator involvement with pT1 to T3 & 5 & 7 & 27 & 14 & 13 \\
Tumour size vs pT & 1 & 1 & 4 & 3 & 6 \\
Depth of invasion $>$10\,mm with pT1 to T2 (AJCC 8th) & 3 & 3 & 3 & 1 & 2 \\
\midrule
\textbf{Any rule} & \textbf{23} & \textbf{22} & \textbf{66} & \textbf{91} & \textbf{112} \\
\bottomrule
\end{tabular}
\end{table}

\subsection{Cascade with escalation}

\begin{figure}[htbp]
\centering
\includegraphics[width=0.68\textwidth]{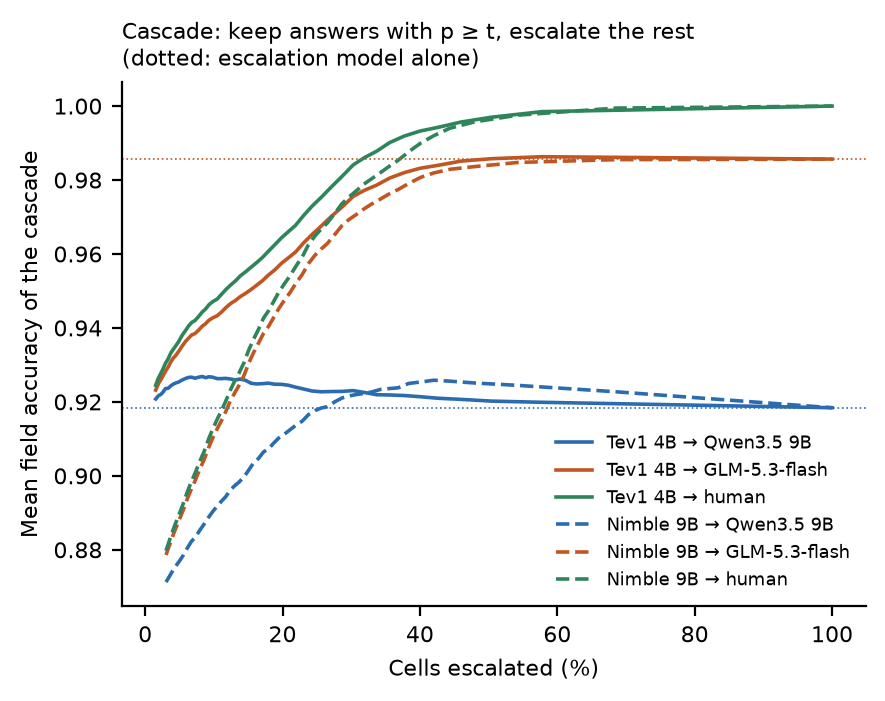}
\caption{Cascade simulation: decision-model answers below a probability threshold are replaced by the answer of Qwen3.5~9B, GLM-5.3-flash or a human reviewer. Mean field accuracy against the share of cells escalated; dotted lines show each escalation model alone.}
\label{fig:cascade}
\end{figure}

When answers below a probability of 0.94 were escalated to GLM-5.3-flash, both decision models came within 0.5 percentage points of GLM alone: Tev1 reached an MFA of 0.982 with 37.7\% of cells escalated, and Nimble 0.981 with 40.1\% (Figure~\ref{fig:cascade}). Escalation to Qwen3.5~9B gave at most 0.927 (Tev1, threshold 0.66, 8.3\% of cells escalated). With a human reviewer checking answers below 0.95, MFA rose to 0.993 for Tev1 with 40.0\% of cells reviewed. At every threshold from 0.8 upwards, however, at least 89\% of reports had at least one escalated cell.

\subsection{Speed and token use}

\begin{figure}[htbp]
\centering
\includegraphics[width=0.72\textwidth]{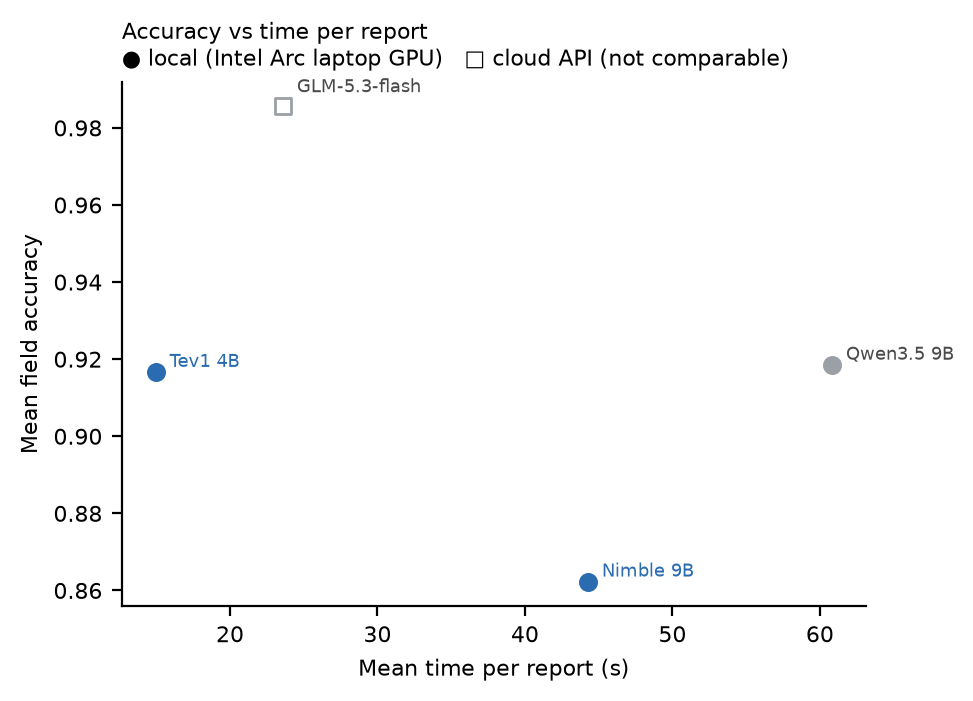}
\caption{Mean field accuracy against mean time per report. Circles: run locally on a GPU. Square: cloud API, not comparable with local times.}
\label{fig:speed}
\end{figure}

On the local GPU, Tev1 took a mean of 15.0\,s per report, Nimble 44.3\,s and Qwen3.5~9B 60.8\,s, so Tev1 was four times faster than Qwen3.5~9B, a generative model with more than twice as many parameters (Figure~\ref{fig:speed}). GLM-5.3-flash took 23.6\,s through the cloud API, which depends on the provider's hardware and is not comparable. Tev1 used a mean of 26,559 input tokens per report and Nimble 146,515 for the same 25 questions. No request failed or exceeded the context limit for either decision model.

\subsection{Robustness}
\label{sec:robustness}

Option order did not explain the decision models' errors. On 200 sampled reports, reversing or shuffling the options, or removing Tx, T0 and Tis, changed pT accuracy by at most 0.02 for either model (all $p \ge 0.12$), and Nimble's wrong answers did not move with the options (Table~\ref{tab:robust}). The only order effect was in Nimble's margin answers (0.850 with the original order, 0.795 reversed; $p < 0.01$), and Nimble over-called close margins in both orders.

The number of questions per request did explain Nimble's pT errors. On the same 200 reports, asking two questions per request instead of 25 raised Nimble's pT accuracy from 0.490 to 0.850 ($p < 0.001$), and from 0.576 to 0.994 among the 170 reports with a recorded pT; most remaining errors were in reports with no recorded pT, where Nimble gave a T category. Nimble's margin accuracy also rose, from 0.785 to 0.850 ($p = 0.002$). Tev1's pT answers were identical with two and 25 questions (accuracy 0.990).

\begin{table}[htbp]
\centering
\caption{Accuracy of pT and margin status on 200 sampled reports, by request size and option order. In brackets: share of answers unchanged from the original order. In the last column only the pT options changed; the margin question was identical.}
\label{tab:robust}
\small
\begin{tabular}{@{}llccccc@{}}
\toprule
Model & Field & \makecell{25 questions\\(main run)} & \makecell{2 questions,\\original order} & Reversed & Shuffled & \makecell{Without\\Tx, T0, Tis} \\
\midrule
Nimble 9B & pT & 0.490 & 0.850 & 0.840 (92.5\%) & 0.830 (95.0\%) & 0.855 (92.5\%) \\
Nimble 9B & Margin & 0.785 & 0.850 & 0.795 (94.5\%) & 0.855 (98.5\%) & 0.860 (99.0\%) \\
Tev1 4B & pT & 0.990 & 0.990 & 1.000 (99.0\%) & 0.995 (99.5\%) & 0.995 (98.5\%) \\
Tev1 4B & Margin & 0.895 & 0.900 & 0.895 (95.5\%) & 0.875 (94.5\%) & 0.885 (98.5\%) \\
\bottomrule
\end{tabular}
\end{table}

Repeated runs gave almost the same answers. Asking all 25 questions again for 100 reports changed 3 of 2,500 answers for Nimble (0.12\%) and 6 of 2,500 for Tev1 (0.24\%). Seven of the nine changes were between a value and ``Not reported'' or ``Not resected''. Nimble's pT accuracy in the repeat was 0.510, so its pT error with 25 questions per request is reproducible.

\section{Discussion}

In a large cohort of oral cavity cancer pathology reports, decision models demonstrated reasonably good performance compared to large language models of a similar or larger size. Specifically, a 4B decision model (Tev1) extracted 25 fields with the same mean accuracy as a 9B generative model run on the same machine, and was four times faster. A second model, Nimble~9B, was less accurate (0.862), and both decision models fell well short of the best generative model, GLM-5.3-flash.

Most decision-model errors were systematic rather than random. They clustered in binned numeric fields, in fields whose correct answer was an absence label such as ``Bone not resected'', and, for Nimble, in pT when many questions were asked in one request. The probability returned with each answer was well calibrated when pooled across fields, and escalating low-probability answers to a stronger model brought both decision models close to that model's accuracy. To our knowledge, this is the first evaluation of decision models on clinical text.

The failure patterns of decision models and generative models were different. Tev1 and Qwen3.5~9B reached the same overall accuracy through different errors. Decision models performed worse for numeric data categorization. The generative models copied the measurement from the report, for example ``1.2\,cm'', and our code converted it to a category. The decision models had to ingest, convert and compare the result with the category limits within one choice, and they answer in a single step without writing out intermediate reasoning \citep{nimble}. Tev1's numeric errors were mostly one category too low for large values; for example, depth of invasion over 10\,mm was called 5 to 10\,mm in 143 reports. Numeric fields should therefore be extracted as values and binned in code, not asked as category questions. This may have implications for the performance of decision models in areas outside healthcare. Qwen3.5~9B errors were different. It tended to perform worse in some categorization questions where information was absent in the reports. The two approaches therefore fail in complementary fields, which favours combining them over choosing one.

Errors were not related to the number of options in a question. Across the 25 fields, the number of options (3 to 10) was not correlated with accuracy (Spearman $\rho = -0.09$ for Tev1, $p = 0.67$; $-0.18$ for Nimble, $p = 0.38$). Errors depended more on the content of the options. Some fields like bone invasion, masticator space and skin required the model to recognise that a structure was not in the specimen or that the report was silent. Errors were more common in these fields. Questions that first ask whether a structure was resected or a measurement was reported, and only then ask for its value, may avoid both errors.

We tested option order because of the pattern of Nimble's pT errors. The pT options were listed as Tx, T0, Tis, T1, T2, T3, T4a, T4b and Not reported, and Nimble's wrong answers for T1, T2 and T3 tumours were Tx, T0 and Tis, each exactly three places earlier in the list. That pattern suggested a positional error, and language models are known to favour options by their position or label in multiple-choice questions \citep{pezeshkpour2024,zheng2023}. The endpoint's rule of resolving ties in favour of the first listed option was a second possible source of a bias toward early options, although exact ties are unlikely with continuous probabilities. The experiments ruled out both explanations: reversing or shuffling the options, or removing Tx, T0 and Tis, left pT accuracy almost unchanged, and the wrong answers did not follow the options to their new positions.

We determined that the cause was the number of questions in each request. With two questions per request instead of 25, Nimble's pT accuracy in reports with a recorded pT rose from 0.576 to 0.994, while Tev1's answers did not change. The token counts revealed a possible cause. Tev1 used about 1,060 input tokens per question, roughly one report and one question, whereas Nimble used about 5,860. In a pilot run with an earlier Ollama release, Nimble needed about 15,600 input tokens per report when the 25 questions were sent one per request, against about 134,000 when they were batched. This is consistent with each of Nimble's per-question prompts carrying the whole question bank, so that the options of other questions compete with those of the question being answered. We did not inspect how the server builds these prompts, so this explanation is inferred. Generative models also lose accuracy when several items share one prompt, especially with long inputs \citep{cheng2023}, and their accuracy depends on where information sits in a long context \citep{liu2024}.

The practical lesson is that request design is part of the model. The same weights gave a pT accuracy of 0.49 or 0.99 depending on how many questions were sent together, so a decision model should be validated in exactly the configuration in which it will be used. The Nimble results reported here are those of the batched configuration, not of the model at its best. Batching also had a cost: Nimble processed 5.5 times as many tokens as Tev1 for the same questions. Because the 8,192-token limit applies to each prompt, the batched configuration also leaves less room for long reports. No request exceeded the limit here, but reports much longer than ours could.

Pooled over fields, the decision models' probabilities were well calibrated (expected calibration error 0.022 and 0.026). This compares favourably with the confidence that generative models state in words, for which expected calibration errors of 0.06 to 0.13 have been reported in medical question answering \citep{boie2026}, and it needs none of the extra computation that uncertainty estimates for generative extraction require \citep{schimmelpfennig2025}. Calibrated token probabilities have been used to accept confident TNM labels from head and neck radiology reports automatically and send the rest for review \citep{gupta2026}. The pooled figures hide field-level failures, however. Tev1 gave its bone-invasion answers a mean probability of 0.845 while being right half the time, because its error was a consistent mapping rather than uncertainty. A probability threshold cannot catch errors of this kind; field-level validation and consistency rules are required.

The cascade results suggest a practical strategy. Escalating whole reports saves no work, because at every threshold from 0.8 upwards at least 89\% of reports had at least one low-probability answer. Escalation should therefore be decided per field and per answer, in four steps:
\begin{enumerate}
  \item During local validation, assign to the decision model only the fields in which it is accurate and well calibrated (for Tev1 in this dataset: PNI, LVI, grade, WPOI-5 and pT), and extract numeric fields as values with a generative model or rules, binning them in code.
  \item Send decision-model answers below a probability threshold to a stronger model, with only the escalated questions; at a threshold of 0.94 this was about 38\% of answers and gave an accuracy of 0.982. The second model must be clearly better than the first: escalating to Qwen3.5~9B gained little (at most 0.927).
  \item Apply the consistency rules to the combined output and send reports that break a rule to a person.
  \item Reserve human review for clinically decisive findings, such as margin status and ENE, when the models disagree or the probability is low.
\end{enumerate}
This is a form of model cascade, in which a cheap model handles most queries and a costly one is called only when needed \citep{chen2023}. Where the stronger model is a cloud service, as GLM-5.3-flash was here, escalated text leaves the hospital, and a strong local model is needed where that is not allowed.

The uses that motivated this study look feasible for some fields and not for others. For triage, sensitivity for PNI, LVI, WPOI-5 and cortical bone invasion was 0.98 to 1.00 with both decision models, so they could flag reports with features that prompt a discussion of adjuvant treatment. They could not be relied on to flag involved margins (Tev1 sensitivity 0.485) or depth of invasion over 10\,mm (0.511). For data quality checking, a decision model run over the reports could flag database entries that disagree with a high-probability answer: at a threshold of 0.95, 60\% of Tev1's answers were accepted at 98.9\% accuracy, so the disagreements in those answers would form a short list to check by hand. Both uses can run on a machine without patient data leaving the hospital, which matters where cloud processing of reports is not permitted.

We did not evaluate Jev or Laya. Jev, from TypeSafe AI, introduced the decision-model category in September 2026, and the Ollama endpoint follows its API \citep{jev}. It is proprietary and offered only as a hosted service, and its size, base model and weights have not been released. Using it would have meant sending pathology reports to an external provider, which routine use inside a hospital should avoid, and a hosted model whose version can change makes results harder to reproduce. We sent de-identified reports to one cloud model, GLM-5.3-flash, only to estimate the upper bound of accuracy, not as a candidate for routine use. Laya, from Convai Innovations, has open weights and runs locally, but it is a 421-million-parameter encoder with a context window of 512 tokens in its English version and 1,024 tokens in its other versions \citep{laya}. Our reports had a median length of 2,102 characters and the longest was about 1,300 tokens, so many would not have fitted in the context together with a question and its options without being truncated or split. Its developers also describe Laya as a base model to be fine-tuned rather than a zero-shot decision model, with accuracy near chance on their own benchmark without fine-tuning. Nimble and Tev1 met our three requirements of open weights, local deployment and a context long enough for a whole report, and both were available through Ollama when the study was run.

\paragraph{Limitations.} This study has several limitations. The reports came from one centre, in one language and reporting style, for one cancer site, and the results may not transfer to other report formats. The reference standard was a clinical database with its own errors: 23 of its records broke at least one consistency rule, and a pathologist review of discordant answers is under way. Numeric fields were scored as categories because decision models cannot return values; this put both approaches on the same scale but tests a coarser task than extracting the measurement itself. The questions were written once, from the generative prompts, and were not tuned; better wording, especially for absence labels, would probably raise accuracy. The main run asked all 25 questions in one request, which understates Nimble's accuracy. Only two decision models were tested, both early releases. The timings come from one machine. Finally, the most accurate comparator ran in the cloud, which may not be acceptable for routine use.

\section{Conclusions}

Decision models extracted presence-or-absence findings from oral cavity pathology reports as accurately as the best generative model, quickly and on local hardware, and returned probabilities that can be used to decide which answers to check. They were unreliable for measurements, for absence labels and, in Nimble's case, when many questions were sent in one request. Used for selected fields, with numeric values extracted by other means and low-probability answers escalated, they may make local, auditable data extraction and quality checking practical. The next steps are to re-run Nimble with fewer questions per request on the full cohort, redesign the questions with absence labels, test the combined pipeline prospectively, and validate it on reports from other centres.

\bibliographystyle{plainnat}
\bibliography{references}

\clearpage
\appendix
\renewcommand{\thetable}{S\arabic{table}}
\setcounter{table}{0}
\section*{Supplementary material}
\addcontentsline{toc}{section}{Supplementary material}

{\small
\begin{longtable}{@{}L{3.6cm}L{5.0cm}L{6.6cm}@{}}
\caption{The nine consistency rules. Each rule was applied to each model's own answers, and to the reference standard, in every report where it applied (results in Table~\ref{tab:consistency}).}\label{tab:rules}\\
\toprule
Rule & Applies when & Counted as a contradiction when \\
\midrule
\endfirsthead
\toprule
Rule & Applies when & Counted as a contradiction when \\
\midrule
\endhead
\bottomrule
\endfoot
Neck dissection side vs right/left node status & Neck dissection side is left, right or bilateral & A side that was not dissected has a node status (Yes or No), or a dissected side is ``Not applicable'' \\ \addlinespace
Positive-node count vs right/left node status & Positive-node count and both side statuses are recorded & 0 positive nodes but either side is ``Yes'', or 1 or more positive nodes but neither side is ``Yes'' \\ \addlinespace
Positive-node count vs pN & pN (N0 to N3b) and positive-node count are recorded & pN0 with positive nodes; pN1 or higher with 0 positive nodes; pN1 with 2 or more positive nodes; pN2b or pN2c with 1 positive node \\ \addlinespace
ENE with no positive nodes & ENE is ``Yes'' and positive-node count is recorded & 0 positive nodes, or no neck dissection \\ \addlinespace
Largest node size vs node count & Largest involved node size and positive-node count are recorded & A node size with 0 positive nodes or no neck dissection, or ``Not applicable'' with positive nodes \\ \addlinespace
``No neck dissection'' vs other node fields & Positive nodes or nodes examined is ``No neck dissection'' & Only one of these two fields is ``No neck dissection'', or a neck dissection side is recorded \\ \addlinespace
Cortical bone, skin or masticator involvement with pT1 to T3 & pT (T1 to T4b) is recorded with cortical bone invasion, skin involvement or masticator space involvement & pT1, pT2 or pT3 \\ \addlinespace
Tumour size vs pT & pT (T1 to T4b) is recorded with tumour size $>$40\,mm, or $\le$20\,mm with depth of invasion $\le$5\,mm & Tumour $>$40\,mm with pT1 or pT2, or tumour $\le$20\,mm with depth $\le$5\,mm and pT2 or pT3 \\ \addlinespace
Depth of invasion $>$10\,mm with pT1 to T2 (AJCC 8th) & pT (T1 to T4b) is recorded, depth of invasion is $>$10\,mm and the AJCC edition is 8th & pT1 or pT2 \\
\end{longtable}
}

{\small
\begin{longtable}{@{}L{2.6cm}L{5.0cm}L{7.6cm}@{}}
\caption{The 25 decision-model questions, with the description sent for each option. Option labels are shown as sent to the models.}\label{tab:questions}\\
\toprule
Field & Question & Options (\textbf{label}: description) \\
\midrule
\endfirsthead
\multicolumn{3}{@{}l}{\emph{Table~\ref{tab:questions} continued}}\\
\toprule
Field & Question & Options (\textbf{label}: description) \\
\midrule
\endhead
\bottomrule
\endfoot
Primary subsite & What is the anatomic subsite of the primary tumor? If there are several tumor foci, give the site of the dominant (largest) tumor. & \textbf{Oral Tongue}: Mobile (anterior two-thirds) tongue. Base of tongue is not oral cavity. \newline \textbf{Floor of Mouth}: Floor of mouth. \newline \textbf{Buccal Mucosa}: Buccal mucosa / cheek. \newline \textbf{Lower Alveolus}: Lower alveolus, lower gingiva or mandibular alveolar ridge. \newline \textbf{Upper Alveolus}: Upper alveolus, upper gingiva or maxillary alveolar ridge. \newline \textbf{Hard Palate}: Hard palate. \newline \textbf{Retromolar trigone}: Retromolar trigone. \newline \textbf{Lip}: Lip. \newline \textbf{Overlapping}: A large tumor extending across several sites without one stated as the primary. \newline \textbf{Not reported}: Site not stated, primary tumor absent, or site is outside the oral cavity (e.g. base of tongue). \\ \midrule
Laterality & What is the side of the primary tumor? Use the side given for the primary tumor specimen, not the side of the neck dissection. If the tumor crosses the midline, give the primary side if reported. & \textbf{Right}: Primary tumor is on the right. \newline \textbf{Left}: Primary tumor is on the left. \newline \textbf{Central}: Report states the tumor is central / midline. \newline \textbf{Bilateral}: Report states the tumor is bilateral. \newline \textbf{Not reported}: Laterality of the primary tumor is not mentioned. \\ \midrule
Focality & How many separate invasive tumor foci are in the primary specimen? Carcinoma in situ, dysplasia and lymph node deposits do not count. A satellite tumor counts only if reported as a distinct primary focus. & \textbf{1}: One focus, or the report does not mention more than one focus. \newline \textbf{2}: Two separate foci, or `multifocal' without a number. \newline \textbf{3}: Three or more separate foci. \\ \midrule
Histological type & What is the histologic type of the carcinoma? Ignore the grade. & \textbf{SCC conventional}: Squamous cell carcinoma, keratinizing or non-keratinizing, any grade, with no variant named. \newline \textbf{verrucous}: Verrucous carcinoma. \newline \textbf{basaloid}: Basaloid squamous cell carcinoma. \newline \textbf{papillary}: Papillary squamous cell carcinoma. \newline \textbf{spindle cell}: Spindle cell (sarcomatoid) carcinoma. \newline \textbf{adenosquamous}: Adenosquamous carcinoma. \newline \textbf{other / not reported}: Any other histology, or histology not stated. \\ \midrule
Grade & What is the grade of differentiation of the carcinoma? When two grades are given (e.g. moderately to poorly differentiated), choose the worse one. & \textbf{Well differentiated}: Well differentiated (grade 1). \newline \textbf{Moderately differentiated}: Moderately differentiated (grade 2), and the worse grade is not poorly. \newline \textbf{Poorly differentiated}: Poorly differentiated (grade 3), including `moderately to poorly'. \newline \textbf{Not reported}: Differentiation is not reported, or there is no primary tumor. \\ \midrule
PNI & Is perineural invasion (PNI) present in the primary tumor? & \textbf{Present}: Report states perineural invasion, perineural involvement or nerve invasion is identified. \newline \textbf{Absent}: Report states perineural invasion is absent, not identified or negative. \newline \textbf{Indeterminate}: Report states perineural invasion is indeterminate. \newline \textbf{Not reported}: Perineural invasion is not mentioned at all. \\ \midrule
LVI & Is lymphovascular invasion (LVI) present in the primary tumor? & \textbf{Present}: Report states lymphatic, vascular, angiolymphatic or lymphovascular invasion is identified. \newline \textbf{Absent}: Report states these are absent, not identified or negative. \newline \textbf{Indeterminate}: Report states lymphovascular invasion is indeterminate. \newline \textbf{Not reported}: Lymphovascular or vascular invasion is not mentioned. \\ \midrule
WPOI & What worst pattern of invasion (WPOI) does the report assign? Use only an explicit WPOI number; do not infer it from words like `infiltrative' or `cohesive'. & \textbf{WPOI5 present}: Report states WPOI-5 (dispersed satellites 1 mm or more apart), without WPOI-4. \newline \textbf{WPOI5 and WPOI4 present}: Report states both WPOI-5 and WPOI-4. \newline \textbf{WPOI5 absent - WPOI4 present}: Report states WPOI-4 without WPOI-5. \newline \textbf{WPOI5 absent}: Report states a worst pattern of 1, 2 or 3. \newline \textbf{Not reported}: Worst pattern of invasion is not mentioned. \\ \midrule
Bone invasion & Is bone involved by the primary tumor? & \textbf{Cortical invasion}: Tumor breaches the cortex and involves bone or the medullary cavity. \newline \textbf{Erosion}: Cortical erosion of bone WITHOUT medullary invasion. \newline \textbf{Not involved}: Bone was clearly resected (e.g. maxillectomy, mandibulectomy or composite resection) and the report states it is not involved or eroded by tumor. \newline \textbf{Bone not resected}: There is no evidence of bone in the specimen at all. \\ \midrule
Masticator space & Does the primary tumor involve the masticator space or its contents (masseter, medial or lateral pterygoid, temporalis)? Do not infer from `skeletal muscle invasion' or `soft tissue' without one of these named. & \textbf{Present}: Report states tumor involves the masticator space or one of these muscles. \newline \textbf{Absent}: Report explicitly states the masticator space or these muscles are not involved. \newline \textbf{Not resected}: The report does not mention the masticator space or these muscles. \\ \midrule
Skin & Does the primary tumor involve the skin of the face? Epidermal involvement with ulceration or nodules counts; dermal involvement without epidermal involvement does not. Mucosa, lip vermilion and soft tissue extension without skin do not count. & \textbf{Present}: Report states the tumor invades or involves facial or overlying skin (epidermis). \newline \textbf{Absent}: Report states skin was resected and is not involved or negative. \newline \textbf{Not reported}: Skin is not mentioned in the report. \\ \midrule
Margin status & What is the final resection margin status for invasive carcinoma? If margins were re-resected, use the re-resection specimen. Margins close to dysplasia or carcinoma in situ, but not invasive tumor, count as clear. & \textbf{Involved}: A margin is involved, or tumor reaches an inked margin. \newline \textbf{Close ($<$ 5mm)}: Any resection margin distance to invasive tumor is less than 5 mm (0.5 cm). \newline \textbf{Clear}: All margins are free and at least 5 mm, or stated as free without distances. \newline \textbf{Not reported}: Margins are not mentioned at all. \\ \midrule
Neck dissection side & Which side(s) of neck dissection were actually submitted? Record the side as submitted, not as planned. & \textbf{Right}: Only a right neck dissection was submitted. \newline \textbf{Left}: Only a left neck dissection was submitted. \newline \textbf{Bilateral}: Both right and left neck dissections were submitted. \newline \textbf{None}: No neck dissection was submitted. \\ \midrule
Right neck nodes & Is metastatic carcinoma present in any lymph node of the RIGHT neck dissection? Take the side from the specimen label (e.g. `right neck level I-III'), not from the side of the primary. & \textbf{Yes}: At least one right-sided lymph node is positive. \newline \textbf{No}: Right-sided nodes were examined and none are positive. \newline \textbf{Not applicable}: There is no right neck dissection. \newline \textbf{Not reported}: A right neck dissection was submitted but node status is not stated. \\ \midrule
Left neck nodes & Is metastatic carcinoma present in any lymph node of the LEFT neck dissection? Take the side from the specimen label, not from the side of the primary. & \textbf{Yes}: At least one left-sided lymph node is positive. \newline \textbf{No}: Left-sided nodes were examined and none are positive. \newline \textbf{Not applicable}: There is no left neck dissection. \newline \textbf{Not reported}: A left neck dissection was submitted but node status is not stated. \\ \midrule
ENE & Is extranodal / extracapsular extension (ENE / ECE) present in any tumor-involved lymph node? & \textbf{Yes}: ENE is present in at least one involved node. \newline \textbf{No}: Report states ENE is absent, or no lymph nodes are involved. \newline \textbf{Not reported}: Nodes are involved but ENE is not mentioned, or ENE is reported as indeterminate. \\ \midrule
pT & What pathological T stage does the report record? Transcribe it, do not derive it (e.g. pT1N0 $\rightarrow$ T1). & \textbf{Tx}: pTx recorded. \newline \textbf{T0}: pT0 recorded. \newline \textbf{Tis}: pTis recorded. \newline \textbf{T1}: pT1 recorded. \newline \textbf{T2}: pT2 recorded. \newline \textbf{T3}: pT3 recorded. \newline \textbf{T4a}: pT4a recorded. \newline \textbf{T4b}: pT4b recorded. \newline \textbf{Not reported}: No T stage is recorded. \\ \midrule
pN & What pathological N stage does the report record? Transcribe it, do not derive it (e.g. pT1N0 $\rightarrow$ N0). & \textbf{Nx}: pNx recorded. \newline \textbf{N0}: pN0 recorded. \newline \textbf{N1}: pN1 recorded. \newline \textbf{N2a}: pN2a recorded. \newline \textbf{N2b}: pN2b recorded. \newline \textbf{N2c}: pN2c recorded. \newline \textbf{N3a}: pN3a recorded. \newline \textbf{N3b}: pN3b recorded. \newline \textbf{Not reported}: No N stage is recorded. \\ \midrule
AJCC edition & Which edition of the AJCC staging system does the report state was used? & \textbf{AJCC6}: AJCC 6th edition. \newline \textbf{AJCC7}: AJCC 7th edition. \newline \textbf{AJCC8}: AJCC 8th edition. \newline \textbf{AJCC9}: AJCC 9th edition / version 9. \newline \textbf{AJCC10}: AJCC 10th edition. \newline \textbf{Not reported}: No staging edition or system is mentioned. \\ \midrule
Tumour size & What is the greatest pathologic dimension of the primary tumor? Use the largest of the stated dimensions (e.g. 2.5 x 1.8 x 0.9 cm $\rightarrow$ 2.5 cm = 25 mm). Do not use specimen dimensions or margin distances. If several foci, use the largest tumor. A value exactly on a cutoff goes in the lower category. & \textbf{$\le$20 mm}: 20 mm (2.0 cm) or less. \newline \textbf{$>$20 to 40 mm}: More than 20 mm and up to 40 mm (2.1 to 4.0 cm). \newline \textbf{$>$40 mm}: More than 40 mm (over 4.0 cm). \newline \textbf{Not reported}: No tumor dimension is stated. \\ \midrule
Depth of invasion & What is the depth of invasion (DOI) of the primary tumor? Use only a value labelled depth of invasion, NOT tumor thickness. A value exactly on a cutoff goes in the lower category. & \textbf{$\le$5 mm}: 5 mm (0.5 cm) or less. \newline \textbf{$>$5 to 10 mm}: More than 5 mm and up to 10 mm (0.6 to 1.0 cm). \newline \textbf{$>$10 mm}: More than 10 mm (over 1.0 cm). \newline \textbf{Not reported}: Depth of invasion is not stated. \\ \midrule
Thickness & What is the tumor thickness? Use only a value labelled tumor thickness, NOT depth of invasion. A value exactly on a cutoff goes in the lower category. & \textbf{$\le$5 mm}: 5 mm (0.5 cm) or less. \newline \textbf{$>$5 to 10 mm}: More than 5 mm and up to 10 mm (0.6 to 1.0 cm). \newline \textbf{$>$10 mm}: More than 10 mm (over 1.0 cm). \newline \textbf{Not reported}: Tumor thickness is not stated. \\ \midrule
Positive nodes & How many lymph nodes are positive for metastatic carcinoma in total, summed across all levels and sides? & \textbf{0}: Nodes were examined and none are positive. \newline \textbf{1}: Exactly one positive node. \newline \textbf{2 or more}: Two or more positive nodes in total. \newline \textbf{Not reported}: A neck dissection was submitted but the number of positive nodes is not stated. \newline \textbf{No neck dissection}: No lymph nodes or neck dissection were submitted. \\ \midrule
Nodes examined & How many lymph nodes were examined in total, summed across all levels and sides? & \textbf{1 to 17}: Between 1 and 17 nodes in total. \newline \textbf{18 or more}: 18 or more nodes in total. \newline \textbf{Not reported}: A neck dissection was submitted but the node count is not stated. \newline \textbf{No neck dissection}: No lymph nodes or neck dissection were submitted. \\ \midrule
Largest involved node & What is the greatest dimension of the largest lymph node involved by metastatic carcinoma? Use the metastatic focus size if stated, otherwise the size of the positive node. Ignore negative nodes and the primary tumor. Sizes may be in mm or cm (21 mm = 2.1 cm). On a cutoff use the lower category. & \textbf{$\le$30 mm}: 30 mm (3.0 cm) or less. \newline \textbf{$>$30 to 60 mm}: More than 30 mm and up to 60 mm (3.1 to 6.0 cm). \newline \textbf{$>$60 mm}: More than 60 mm (over 6.0 cm). \newline \textbf{Not reported}: There are positive nodes but no size is stated. \newline \textbf{Not applicable}: There are no positive nodes, or no neck dissection. \\
\end{longtable}
}

\end{document}